\documentclass{article}

\usepackage{microtype}
\usepackage{graphicx}
\usepackage{booktabs} % for professional tables

\usepackage{xspace}
\newcommand*{\eg}{e.g.\@\xspace}
\newcommand*{\ie}{i.e.\@\xspace}
\usepackage{bm}
\usepackage{amsmath}
\usepackage{amssymb}
\usepackage{subfigure}
\usepackage{color,soul}
\DeclareMathOperator*{\argmax}{argmax}

\usepackage{hyperref}

\usepackage[accepted]{icml2019}

\usepackage{stfloats}

\icmltitlerunning{ICML 2019 Workshop on Uncertainty and Robustness in Deep Learning}

\begin{document}

\twocolumn[
%\icmltitle{Uncertainty Distillation: Simple, Scalable and Efficient Uncertainty Estimates}
\icmltitle{Efficient Evaluation-Time Uncertainty Estimation by Improved Distillation}



\icmlsetsymbol{equal}{*}

\begin{icmlauthorlist}
\icmlauthor{Erik Englesson}{kth}
\icmlauthor{Hossein Azizpour}{kth}
\end{icmlauthorlist}

%\icmlaffiliation{kth}{Division of Robotics, Perception, and Learning, School of Electrical Engineering and Computer Science, KTH (Royal Institute of Technology), Stockholm, Sweden}
\icmlaffiliation{kth}{RPL, KTH, Stockholm, Sweden}

\icmlcorrespondingauthor{Erik Englesson}{engless@kth.se}
%\icmlcorrespondingauthor{Hossein Azizpour}{azizpour@kth.se}

\icmlkeywords{Machine Learning, ICML}

\vskip 0.3in
]



\printAffiliationsAndNotice{}  % leave blank if no need to mention equal contribution

\begin{abstract}
% We propose a novel distillation procedure to obtain computationally-efficient uncertainty estimates at evaluation time. The procedure has several novelties helping its robustness to in and out-of-distribution uncertainties. Using the proposed procedure, we achieve state-of-the-art predictive uncertainty estimates efficiently. Additionally, our analysis sheds light on the design of the standard distillation technique through which we improve upon it.
%uncertainty is good
% for them to be practical in real-world applications, they need to be fast to compute and have low memory footprint
% we propose an elaborate class-posterior amortization framework which is agnostic to the original training procedure and achieve results on-par to the original model using state-of-the-art architectures, various uncertainty measures and on multiple datasets.

In this work we aim to obtain computationally-efficient uncertainty estimates with deep networks. For this, we propose a modified knowledge distillation procedure that achieves state-of-the-art uncertainty estimates both for in and out-of-distribution samples. Our contributions include a) demonstrating and adapting to distillation's regularization effect b) proposing a novel target teacher distribution c) a simple augmentation procedure to improve out-of-distribution uncertainty estimates d) shedding light on the distillation procedure through comprehensive set of experiments.

\end{abstract}

\section{Introduction}
\label{sec:intro}
% why uncertainty is important
Deep neural networks are increasingly used in real-world applications thanks to their impressive accuracy. Nevertheless, many of these applications involve human users which necessitate high level of transparency and trust beside the accuracy. A crucial ingredient to enable trust, is to associate the automatic decision with a calibrated uncertainty.\\
% current methods of uncertainty very briefly
Different techniques have been developed to obtain uncertainty estimation from deep networks including Bayesian modeling using variational approximation~\cite{graves2011nips}, expectation propagation~\cite{hernandez2015icml}, sampling~\cite{gong2018arxiv} and non-Bayesian methods such as bootstrapping~\cite{lakshminarayanan2017nips} and classification margin~\cite{geifman2017nips}.\\
% the Bayesian approximation class is slow in both training and inference and do not perform best
While many of those models achieve acceptable uncertainty estimates~\cite{lakshminarayanan2017nips}, they are notoriously slow to train and evaluate which makes them non-viable for real-world implementation.
% there are methods that make the training much more manageable short
Methods have been developed to expedite the training process. \cite{gal2016icml, teye2018icml} cast standard deep networks as approximate Bayesian inference, \cite{welling2011icml} use Langevin dynamics in tandem with stochastic gradient descent, and \cite{huang2017iclr} employ the training snapshots to avoid sampling and/or multiple training runs. \\
% to employ uncertainty in real-world application we need faster and more computationally and memory-wise efficient models
% those models are not efficient in both
All these techniques effectively improve the computational complexity of training, but remain prohibitively expensive at \textit{evaluation}, essentially due to the multiple inference required for uncertainty estimates at test time. This is despite the fact that \textit{computational and memory-footprint is of high concern for real-world applications} where the evaluation model needs to be deployed in products with limited computational and memory capacity.\\
% the focus of this work is to do this by devising the best way to achieve simple and efficient uncertainty at evaluation
% amortization
The focus of this work is to address this issue. We devise an algorithm to efficiently obtain uncertainty estimates at evaluation time \textit{irrespective of the modelling choice}. Common deep networks for epistemic uncertainty produces either samples of the posterior $P(\bm{\theta}|D)$ or a parametric approximation of it $q_{\bm{\omega}}(\bm{\theta})$, where $\bm{\theta}\in\Theta$ is the model parameters (\ie weights and biases of the network), $D=\{(\bm{x}_i,y_i)\}_{i:1...N}$ the training data, and $\bm{\omega}$ the parameters of the approximating distribution. 
To marginalize over the parameter uncertainty, usually, $m$ samples of the parameters $\{\bm{\theta}_1,...,\bm{\theta}_m\}$ are obtained \eg through bootstrapping~\cite{lakshminarayanan2017nips} or sampling of $q_{\bm{\omega}}(\bm{\theta})~$\cite{gal2016icml}:
%Then, for evaluating a new $\bm{x}$ using $m$ samples of the true posterior $\{\bm{\theta}_1,...,\bm{\theta}_m\}$ coming either from direct bootstrapping ~\cite{lakshminarayanan2017nips} or sampling of $q_{\bm{\omega}}(\bm{\theta})~$\cite{gal2016icml}. Marginalizing out the parameters, the class-posterior can be approximated by:
\begin{align}
P(y|\bm{x},D) \approx \frac{1}{m}\sum_{i=1}^{m}P(y|\bm{x},\bm{\theta}_i) \quad \{\bm{\theta}_i\} \sim P(\bm{\theta}|D).
\label{eq:target}
\end{align}
This summation is the source of evaluation memory/time complexity. 
In this work, we aim to train a single deep network with parameters $\bm{\theta'}$ potentially from a different parameter space $\bm{\Theta'}$ that minimizes a divergence from the mean distribution of Eq~\ref{eq:target} for all $\bm{x}$. The teacher-student setup of \cite{hinton2015nips} is suitable for this purpose.
%In this work, we aim to find $\bm{\theta'}^*$, \textit{potentially from a different parameter space}, that amortizes this integration through a single evaluation. Thus, we find a $\bm{\theta'}$ that minimizes divergence from the true class-posterior $\smash{\displaystyle\min_{\bm{\theta}'}}\, \text{D}(P(y|\bm{x},\bm{\theta}_1,...,\bm{\theta}_m),P(y|\bm{x},\bm{\theta'}))$. For this, the teacher-student setup of \cite{hinton2015nips} is suitable.

\section{Method}
\label{sec:method}
Our goal is to optimize the parameters of a single (student) network $\bm{\theta'}$ to produce class-posterior similar to our target (teacher) distribution \textit{for both in and out-of-distribution samples}. One measure is the KL-divergence:
\begin{align}
\min_{\bm{\theta}'} \sum_{\bm{x}\in X}\text{D}_{KL}(P(y|\bm{x},D)\|P(y|\bm{x},\bm{\theta'}))
\label{eq:goal}
\end{align}
%These parameters are found by the minimization of Equation \ref{eq:goal}, which is equivalent to minimizing the standard cross-entropy loss using the distributions in Equation \ref{eq:target} as targets.
with $X$ being \textit{student} training set, and $P(y|\bm{x},D)$ is approximated as in Eq~\ref{eq:target}. We denote this training procedure as \textit{vanilla distillation}. Note that Eq~\ref{eq:goal} exclude the hyperparameters of the standard distillation  \cite{hinton2015nips}: temperature $T$ and mixing parameter $\alpha$.
\subsection{Target distribution as regularization}
The student needs to learn a dispersed distribution as target in contrast to the teacher's hard label (\ie Kronecker delta pmf, $\delta_{y,y_i}$), which we assume is a ``more difficult" task that requires higher model capacity. Indeed, we have empirically observed that students, with the same architecture as the teacher, tend to converge slower and are less prone to overfitting. Furthermore, \cite{hinton2015nips} and \cite{balan2015nips} noticed a lower $L2$ regularization weight is needed for the student. Based on these observations, we propose the following two modifications to standard distillation. 

\textbf{Higher capacity students.} One way to address this phenomenon is to increase the student's capacity (w.r.t. teacher) to account for the additional complexity. That is, we assume $|\bm{\Theta'}|>|\bm{\Theta}|$. This can be done, for instance, by increasing the depth or width of the student network. \footnote{Although this results into a less efficient student, it will still be eminently more efficient than evaluating multiple teachers. Also, the student, after being fully trained, can be compressed using various existing approaches (\eg\cite{zhuang2018nips}).}

\textbf{Sharper target distribution.} %Alternatively, we can decrease the target distribution's entropy. We obtain the sharpened distribution q from the teacher distribution p by:
Alternatively, for each sample, we can decrease target's entropy H($\cdot$) by sharpening the teacher's distribution $p$ to a new target distribution $q$:
% \begin{align}
%     \label{eq:sharp_target}
%     q = (1-\alpha)p + \alpha r \quad {\scriptstyle \text{with} \,\, \text{H}(r) < \text{H}(p) \text{ and } 0<\alpha<1}.
% \end{align}
\begin{align}
    \label{eq:sharp_target}
    &q = (1-\alpha)p + \alpha r \quad \\
    &{\scriptstyle\text{with}} \,\, \text{H}(r) < \text{H}(p) ,\,\, \argmax_y{r}=\argmax_y{p}, \,\, 0<\alpha<1. \nonumber
\end{align}
$r=\delta_{y,y_i}$ gives the formulation of knowledge distillation without temperature $T$. This can explain the improvement \cite{hinton2015nips} observed by adding the true labels to the distribution for the \textit{correctly classified examples}. %, which reduces the entropy of the \textit{correctly classified examples}.
%\vspace{-0.6cm}
\subsection{Proper class-posterior distribution}
Here we pose the question of \textit{whether to follow the teacher even when it makes wrong decisions} (\ie $y_{max}\neq y_i$, where  $y_{max}\triangleq \argmax_y p_i(y)$). From the perspective of predictive uncertainty, we argue that it is only reasonable for a target distribution to be as ``faulty" as a uniform distribution. However, a uniform distribution loses the dark knowledge in the wrong prediction, so we propose the following alternative distribution for each misclassified sample $i$:
\begin{align}
    \label{eq:proper_target}
    q_i = (1-\alpha_i)p_i + \alpha_i \delta_{y,y_i} \,
    {\scriptstyle \text{with} \, \frac{p_i(y_{max})-p_i(y_i)}{p_i(y_{max})-p_i(y_i)+1}<\alpha_i\leq1}.   % \nonumber
\end{align}
$\alpha$ at its minimum makes a new target distribution $q$ with maximum mass on the correct class while still retaining maximal mass on non-correct classes and thereby dark knowledge. This complements the previous argument by explaining the observed improvement in \cite{hinton2015nips} of mixing in the true labels for the \textit{wrongly classified examples}.
%This proposed proper target offers another reason why the target in \cite{hinton2015nips} can make sense (for wrongly classified samples by the teacher).

% FOR NIPS: this most likely doesn't work because on the training set the number of wrongly classified samples by the teacher is very low

% FOR NIPS: actually if the capacity is large enough such that there no reason to sharpen the distributions (of course except for the case that the teacher is wrong) then we might want to actually disperse the distribution along the lines of label smoothing and penalizing the overconfident predictions. Here would probably be where class-specific distributions come in (because they are better than the uniform??!!)
\vspace{-0.3cm}
\subsection{Robustness to out-of-distribution (OOD) samples}
\label{sec:method:ood}
Ideally we want the model to have high uncertainty when it is presented with samples that are out of the training distribution, \ie, $O=\{\bm{x}|P(\bm{x},y)\approx0\,\,\forall y\in\mathcal{Y}\}$.\footnote{Note that, marginalizing over $y\in \mathcal{Y}$ does not give $P(\bm{x})$ since $\mathcal{Y}$ is limited to the current task excluding a ``negative'' class.} We posit that this set includes two important subsets.

\textbf{Natural set.} OOD samples can come from the support of $P(x)$, \ie, $O_{natural}=\{\bm{x}|P(\bm{x})>\epsilon\}\cap O$. For instance, an image of a car is a natural OOD sample for a cat vs dog classification task. 
%To make the student robust to these samples we propose to sample from the natural image manifold during training the student. 
\cite{li2018arxiv} uses a large unlabeled student training dataset for this purpose. %Alternatively, one can use generative models.

\textbf{Unnatural set.} Unnatural OOD samples come from the rest of the space, \ie, $O_{unnatural}=\{\bm{x}|P(\bm{x})\approx0\}$ which are important for defending against adversarial attacks. \cite{lakshminarayanan2017nips} uses adversarial training to become robust to this set of OOD samples.

% NIPS: Primarily, we want to respect the support of P(x), that is to assign non-uniform distribution only to samples coming from this support. That actually means that adversarial examples (by adding noise to the samples of manifold) should quickly approach a uniform predictive distribution. However, to keep the smoothness, at the same time, we would like to push for natural samples to have non-uniform probabilities using data augmentation, generative models, etc.

\vspace{-0.3cm}
\section{Contributions}
\label{sec:contrib}
The contributions of this work can be summarized as:
\begin{itemize}
\vspace{-0.35cm}
    % we recognize the regularization effect of teacher-student learning and optimize the hyper-parameters (including architecture) of the student and increase the capacity of the student
    \item we recognize the regularization effect of dispersed target distributions and accordingly suggest techniques to improve the distillation process
    \vspace{-0.1cm}
    % understanding the formulation of KD
    \item we provide justification for the particular target distribution of standard distillation in \cite{hinton2015nips}
    %novel targets addressing the issue with the standard distillation
    %\item following same line of reasoning, we design novel target distribution that improves the distillation
    % novel adversarial training for better out-of-distribution distillation
    %\vspace{-0.1cm}
    \item we propose a simple and yet effective technique for distillation of out-of-distribution predictive uncertainty
    % using the procedure above we demonstrate state-of-the-art performance using various quantitative measures.
    \item we conduct a comprehensive set of experiments and evaluations to study the aforementioned aspects
    %\vspace{-0.1cm}
    % should we talk about getting rid of KD hyperparametrs to optimize (temperature and alpha)?
    % \item high-quality results are achieved by relieving the need for the hyperparameters of the standard distillation ($\alpha, T$).
\end{itemize}

\vspace{-0.5cm}
\section{Related Work}
\label{sec:related}
% we skip all these works due to space limit for the workshop paper up until the line
% general intro on uncertainty estimation with deep networks

% different families, approximate Bayesian methods, non-Bayesian methods

% variational-inference based
% sampling-based
% others (moment matching, expectation propagation, etc.)

% simple&effective, calibration, etc.

% methods that try to make training more efficient (snapshot, SGLD, MCDO/MCBN)

%%%%%%%%%%%%%%%%%%%%%%%%%%%%%%%%%%%%%%%%%%%%%%%%%%%%%%%%%%%%%%%%%%%%%%%%%%%%%%%%%%%%%%%%%%%%%%%%%
% methods that try to make evaluation fast and our improvement upon all of them.
Here we briefly present the recent works that address the computational efficiency of evaluating predictive uncertainty and delineate our work with respect to them. \\
% Knowledge distillation
\cite{hinton2015nips} coined knowledge distillation to summarize an ensemble. They focused on the accuracy of the student and not the uncertainty estimates. Our work sheds light on their design choices and is more elaborately designed for the purpose of uncertainty distillation.\\
\cite{li2018arxiv, gurau2018arxiv, balan2015nips} are the closest to our work, they use \cite{hinton2015nips} to distill ensemble networks, Monte Carlo sampling of dropout networks \cite{gal2016icml} and approximate posterior samples of SGLD \cite{welling2011icml} respectively. They use  distillation in its standard form, 
%thus our technical novelties and theoretical observations are complementary to those works.
thus our observations and proposed modifications are complementary to those works.
\cite{li2018arxiv} addresses the problem of OOD prediction with an unlabeled dataset, whereas we propose different and potentially complementary procedures.\\
% Uncertainty Estimates and Multi-Hypotheses Networks for Optical Flow (FOR WORKSHOP I'd SKIP COMPARING THEM SINCE THEY'RE TANGENTIALLY RELATED AND WE NEED SPACE!
% \cite{ilg2018eccv}\\
% Large Scale Distributed Neural Network Training Through Online Distillation
\cite{anil2018iclr} designs a technique to distill ensembles in an online fashion, focused on a distributed training scenario. Their goal is to match and improve the accuracy as opposed to predictive uncertainty.\\
% Reducing Overconfident Errors Outside The Known Distribution
% Dropout Distillation for Efficiently Estimating Model Confidence
% The concurrent works of \cite{li2018arxiv, gurau2018arxiv} attack the same problem. Apart from the additional technical novelty we bring, most important difference of our work from the high level is that, we acknowledge that the class-posterior amortization is a different task than the original inference and thus requires an elaborate design with a higher-capacity architecture. Doing so, we achieve significantly improved results.
% maybe also comment on hyperparameters of normal distillation which is removed from our work (when alpha=0) also we don't see any use for temperature scaling
%Deterministic VI( ICLR 2019) (While formally elegant, our method is better in practice since it can mimic any high-quality training and it achieves better results)
Finally, \cite{wu2019iclr} proposes a method to deterministically propagate uncertainty of model parameters and activations to the output layer. While this elegant approach circumvents the computational burden of sampling the parameter posterior, it achieves inferior results compared to the ensemble model of \cite{lakshminarayanan2017nips}. %On the other hand, we improve the results at the same time as making the evaluation efficient. 
% Maybe: Born-again, Label-smoothing, Penalizing confident output predictions

\vspace{-0.3cm}
\section{Experiments}
\label{experiments}

\begin{figure*}[t!]%
    \centering
    \subfigure[In-Distribution]{\label{fig:vanilla:in}\includegraphics[width=.3\linewidth]{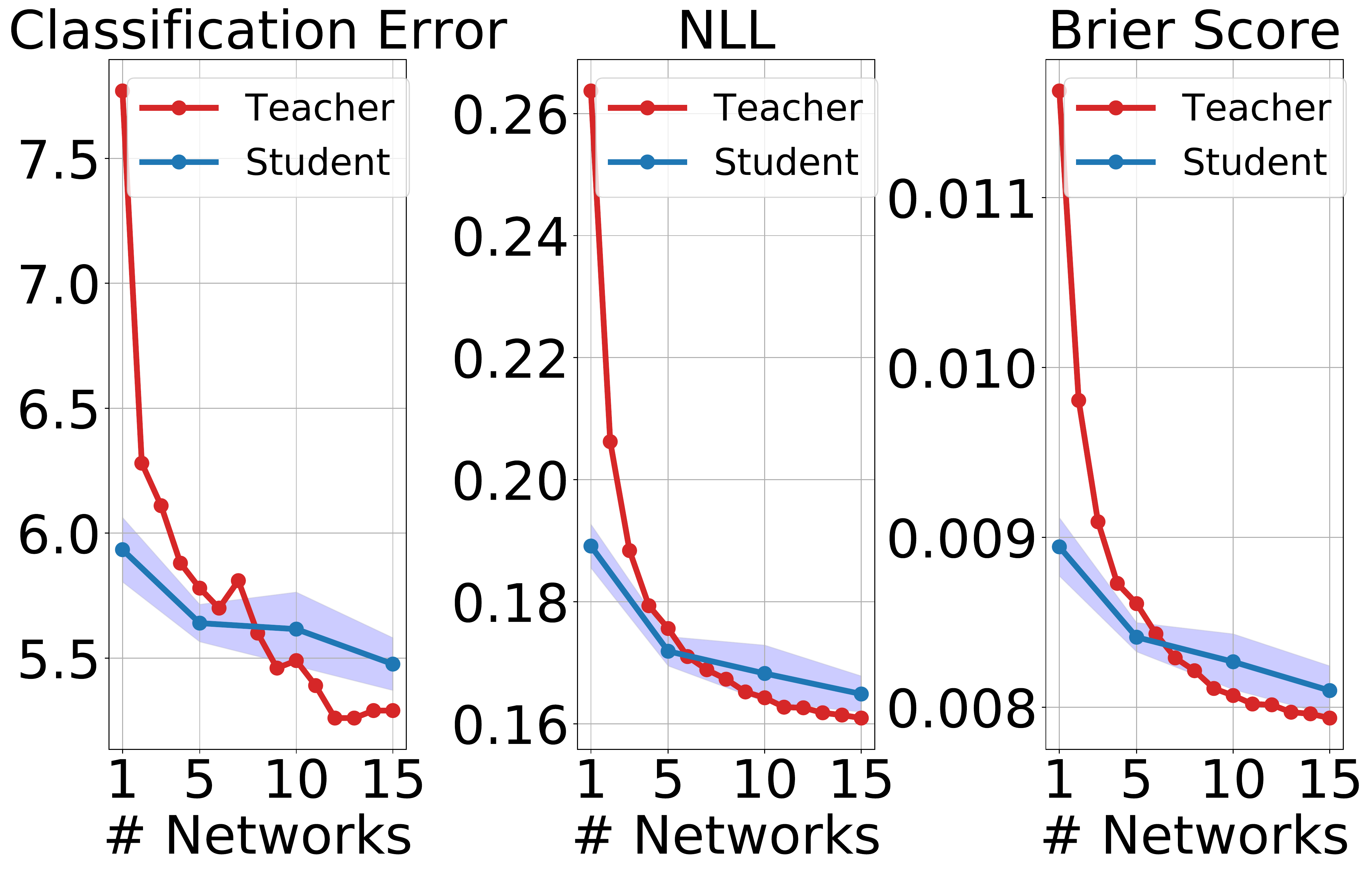}}
    \qquad
    \subfigure[Out-of-Distribution]{\label{fig:vanilla:out}{\includegraphics[width=.3\linewidth]{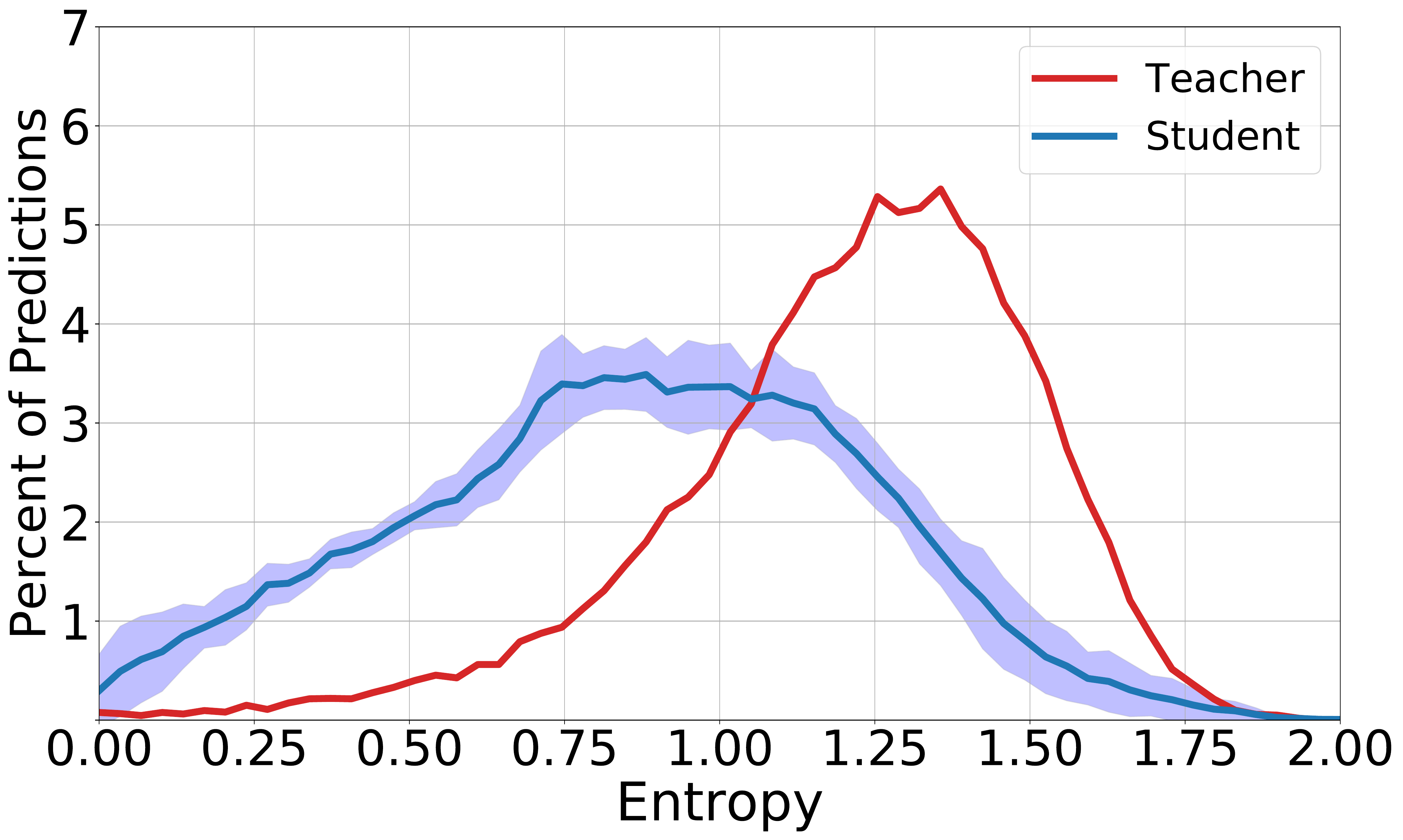} }}%
    \vspace{-0.4cm}
    \caption{\textbf{Evaluation of the predictive uncertainty for a) in-distribution and b) out-of-distribution samples}. Models are trained on CIFAR-10 with vanilla distillation (standard distillation with $\alpha=0, T=1$). For the in-distribution plot, we vary the ensemble size. The student is trained on a teacher with the corresponding number of networks. Student and teacher networks have the same capacity (depth 9). For out-of-distribution plot, we evaluate the models on SVHN and create a histogram over the entropy of the predicted distributions. The quality of the predictive uncertainty is decent for the in-distribution, but there is room for improvement for out-of-distribution samples.}%
    \vspace{-0.3cm}
    \label{fig:vanilla}%
\end{figure*}
% Two sentences and move rest to appendix.
% Strategy: 1) Bullet points for things I want to write about, 2) Quickly write about it
% Only write about results that are currently in the paper. Don't care to much about how everything is connected. It can change.

% What is most important here. Evaluation metrics, things that are common over all runs, datasets are maybe not important here?
We use the state-of-the-art ensemble technique proposed in~\cite{lakshminarayanan2017nips} as the teacher. We measure calibration of in-distribution predictive uncertainty through negative log-likelihood (NLL) and Brier score.
%, and calibration plots
We evaluate the robustness to OOD samples via entropy histograms. The experimental results of the student is reported as the mean and std of 5 runs unless stated otherwise. We use CIFAR10 as the main dataset, and report some results on MNIST and CIFAR100. See appendix Sec~\ref{sec:app:exp_details} for details. 

% 2.0
\subsection{Vanilla distillation} % Or Optimization depending on how much we add
% KL-1 vs KL-2 etc -- need to generate this, low prio
% TEXT: In the first experiment, we investigate which measure is best to use in the minimization. We consider the two directions of KL-divergence and the Jensen-Shannon divergence. The results for varying epochs are shown in Table \ref{}.

% vanilla distillation: alpha=0, no augmentation, same student-teacher capacity: here we show that results look generally good for in-distribution and slightly worse for out-of-distributions
% What are the points I want to get across here?
% Vanilla distillation work pretty well for in-distribution out of the box. Out-of-distribution is a different question. 
% Why? To motivate the work of doing something about OOD? What is the correlation to in-distribution? In distribution we just want to understand why increasing alpha make sense? Or connect to regularization.
First, we show that vanilla distillation produces decent predictive uncertainty for the in-distribution samples, while it is significantly worse on the OOD samples, see Fig~\ref{fig:vanilla}. Results for other network depths are in Fig~\ref{fig:varying-num-networks} in the appendix. %Ideally, we would have out-of-distribution predictions that are as close to uniform as possible (entropy $\sim$ 2.3 for 10 classes), but this is not achieved by the ensemble. 
In Fig~\ref{fig:vanilla:out}, we can see that the student is more over-confident in its predictions compared to the teacher. It is still interesting that this simple baseline without hyperparameters performs on-par with the ensemble teacher. We now further improve upon these results using our proposed techniques.

% Here: 
% 1. Could either make this like the overview section where we motivate our problem, i.e. shown with the figure what the problem is and then. In section "Target.." we investigate how to improve the performance of in-distribution samples. In section "Robustness to domain shift" we show our improvements on OOD. Maybe that figure should actually come much higher, before the method section?
% -- Isssue, where does KL thingy come in then.. haven't looked at the results of these runs..
% 2. Feels a bit out of place if just mention that in distribution works well but not ood?

% NLL vs validation cost -- need to generate this or just mention,
%TODO: Add NLL vs validation cost figure, if there is time. Argue that if we can improve the performance on the in-distribution by modifying the targets from the teachers, then why not.

\subsection{Target distribution as regularization}
% Main goal is to experimentally show that the target distribution acts as a regularizer. 
%1. Not overfitting to NLL
%2. Train for longer: Mention amount of training for student vs teacher
%3. Generalization gap, look at teacher NLL vs vanilla student trained on single teacher? Column 1. Look at first column of Figure 2 b,c. (diff in epochs though but teacher would have overfitted.. running this, also compare to the best NLL of teacher in this longer epoch case)
%4. Varying capacity, i.e. b) and c)
%5. Alpha
% Is the best performance always gotten from harder labels? Less wrong predictions?

% How does the number of networks or capacity of the teacher affect the performance/regularization of the student? See that lower num and capacity gives better results for student than the teacher, but in terms of the total NLL it is still lower than the others right? Maybe an additional plot with NLL for students would be good too. Is there a difference between capacity and num networks? Is it easier to mimic an increase in depth vs number of networks? 
% Increasing the capacity gives better results?

% Side-note: Once again notice that a=0 works quite well for in-distribution- especially when increasing the capacity of the student. Discuss that we still get a lot of memory savings.

% 1 & 2
We have observed that the teachers quickly overfit to NLL after the first drop in learning rate (while accuracy still improves~\cite{guo2017icml}); this behavior, however, is not observed for the students. Furthermore, the convergence time of the students is far longer than the teachers -- 2500 vs 85 epochs. These observations hint that the student learning process is more regularized than its teacher's counterpart. In the following, we take measures based on this observation.

\begin{figure*}%
    \centering
    \subfigure[Teachers]{\label{fig:a0:teacher}{\includegraphics[width=.3\linewidth, trim={0 0 0 0},clip]{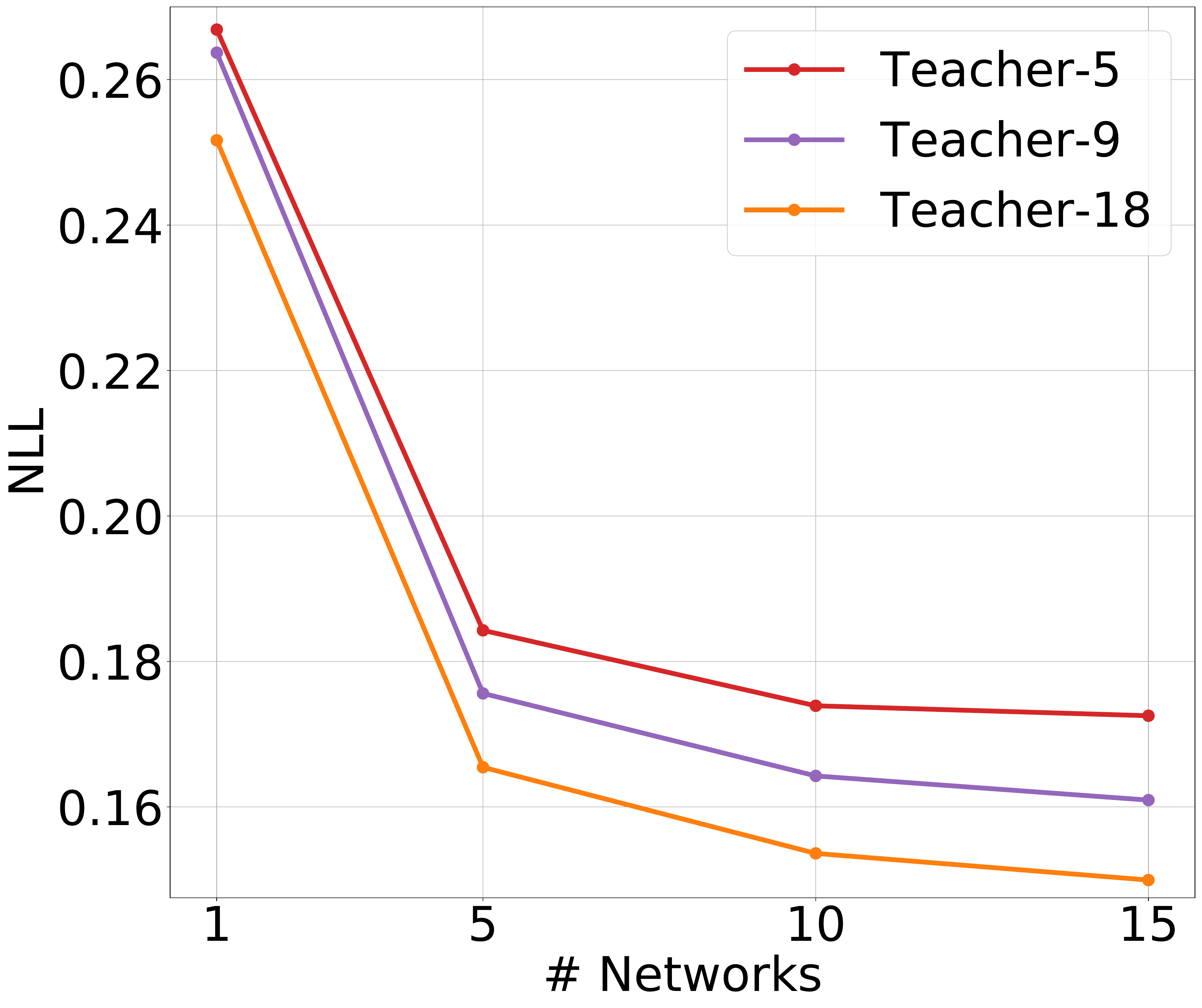}}}%
    \subfigure[Varying Student Depth]{\label{fig:a0:varying-student-depth}{\includegraphics[width=.3\linewidth]{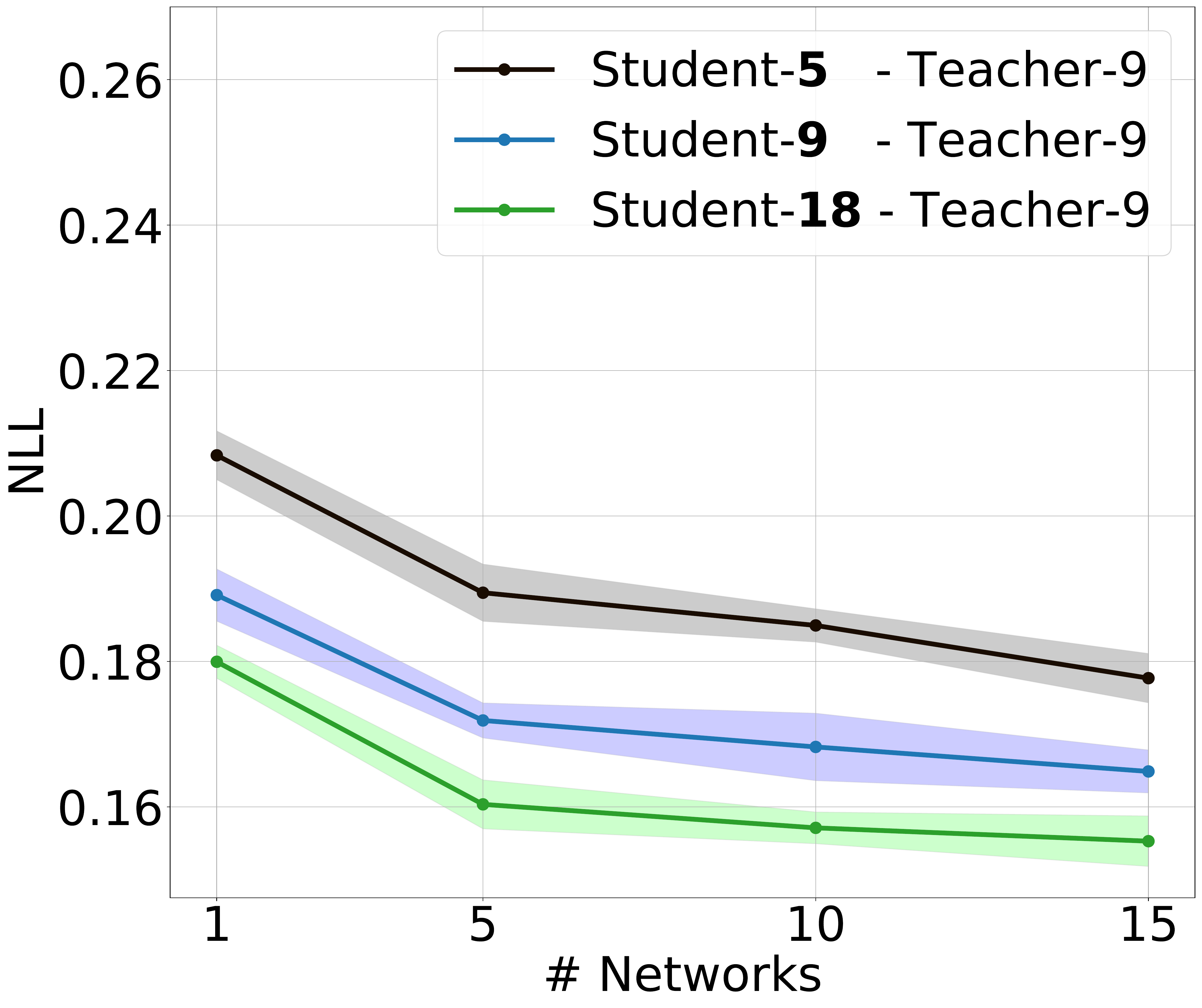}}}%
    \subfigure[Varying Teacher Depth]{\label{fig:a0:varying-teacher-detph}{\includegraphics[width=.3\linewidth]{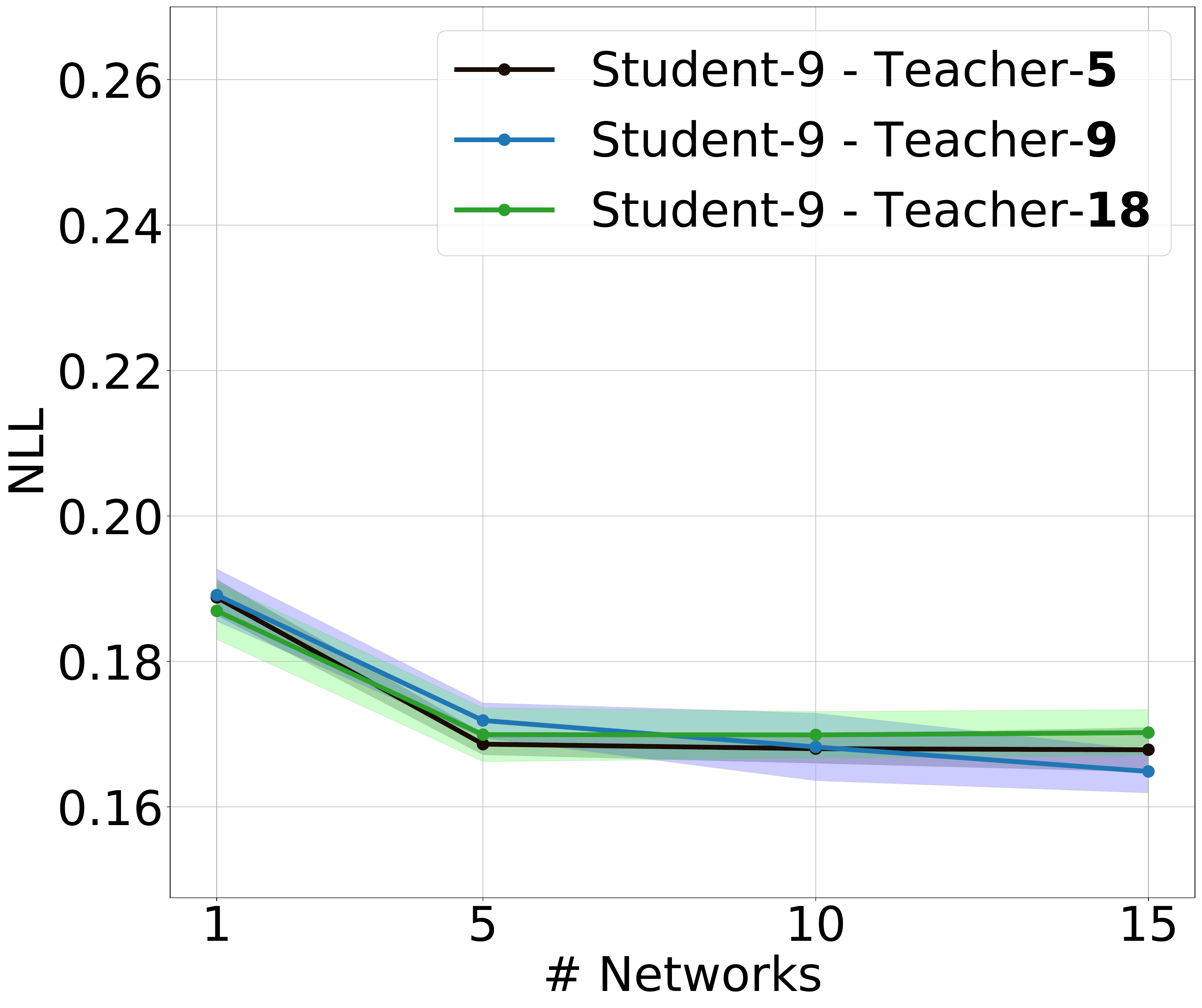}}}%
    %\subfloat[CIFAR100-SVHN]{{\includegraphics[width=.2\linewidth]{student-1-varing-capacity-networks.pdf}}}%
    %\hspace*{0em}
    %\vspace{-0.3cm}
    \caption{\textbf{Capacity and ensemble size of teachers vs corresponding students on CIFAR-10:} Figure (a) shows how the performance of the teacher depends on the number of networks and the capacity of each network. The performance of the student is consistently improved by increased depth(b), while the depth of the networks in the teacher does not significantly affect the student(c). All students are trained using vanilla distillation. }%
    \vspace{-0.5cm}
    \label{fig:a0}%
\end{figure*}

% Introduce experiment for 3, 4.
%% NIPS: The following two sentences need evidence from how peaky the distributions become as you increase the number of networks in the ensemble and the depth of the teachers. Then, this peakiness should make the student's job *easier* i.e. less regularized. We should be able to show that in convergence time and/or the necessity for higher-capacity students, or other experiments.
%First, we show how different targets regularizes the student in the vanilla distillation setting. To do this, we create several different targets by varying the capacity and the number of networks of the ensemble. 
\textbf{Higher capacity students.} Fig~\ref{fig:a0:teacher} serves as a baseline for how the teacher performs for varying number of networks and network depths. In Fig~\ref{fig:a0:varying-student-depth} we consistently observe better NLL as the student's depth is increased. The results for other teacher depths are shown in Fig~\ref{fig:appendix:a0-varying-student-depth} in the appendix. More interestingly,
%increasing the capacity of a single network when using hard targets have little effect on the NLL, however, increasing the capacity of the student has a much bigger effect. .To more clearly observe the effect of increasing the depth of the student we look at Fig~\ref{fig:varying-student-depth}, \ref{fig:vary-student-depth-teachers-curves} in the appendix.
increasing the student's capacity is more effective than increasing the teacher's, see Fig~\ref{fig:a0:varying-teacher-detph}. This can be due to the same regularization effect. This was also observed for students of depth 5 and 18, see appendix Fig~\ref{fig:appendix:a0-varying-teacher-depth}.

%In Fig~\ref{fig:a0:teacher} in the appendix, as we move from bottom-left to top-right (\ie deeper and larger ensemble), the teacher targets exhibit lower entropy (by proxy lower teacher NLL). Fig~\ref{fig:a0:sd5},\ref{fig:a0:sd18} show corresponding students of depth of 5 and 18. We can interestingly see that the difference between the bottom-left to top-right point of student with depth 5 and 18 \textit{decreases}.
%Fig~\ref{fig:a0-diff} shows the NLL difference between the target generating teacher and its corresponding student. We observe that the student matches the performance of the teacher.\\
Finally, all the figures crucially indicate that the improvement in student performance by increased depth is not merely because the original task demanded larger networks. That can be seen by comparing the improvements in the ensemble performance to student's as the depth increases.

\textbf{Sharpening the target distribution.} Another way to address the regularization of dispersed target distributions is to lower the entropy as in \cite{hinton2015nips}. Interestingly, we empirically observed the effect of $\alpha$ diminishes as the capacity of the student is increased. Appendix Tab \ref{tab:sharpen} shows that a student of depth 18 trained using a teacher of depth 5, does not significantly benefit from an increase in $\alpha$. 
For results of sharpened targets on MNIST and CIFAR100, see Appendix Fig \ref{fig:appendix:mnist} and Tab \ref{table:appendix:cifar100-sharpen}, respectively.

\begin{figure*}[b!]%
    \vspace{-0.3cm}
    \centering
    % \subfigure[MNIST-CIFAR10]{\label{fig:ood:mnist}{\includegraphics[width=.3\linewidth]{Figures/mnist-out-of-distrib.pdf}}}%
    % \qquad
    \subfigure[CIFAR10-SVHN]{\label{fig:ood:cifar10}{\includegraphics[width=.3\linewidth]{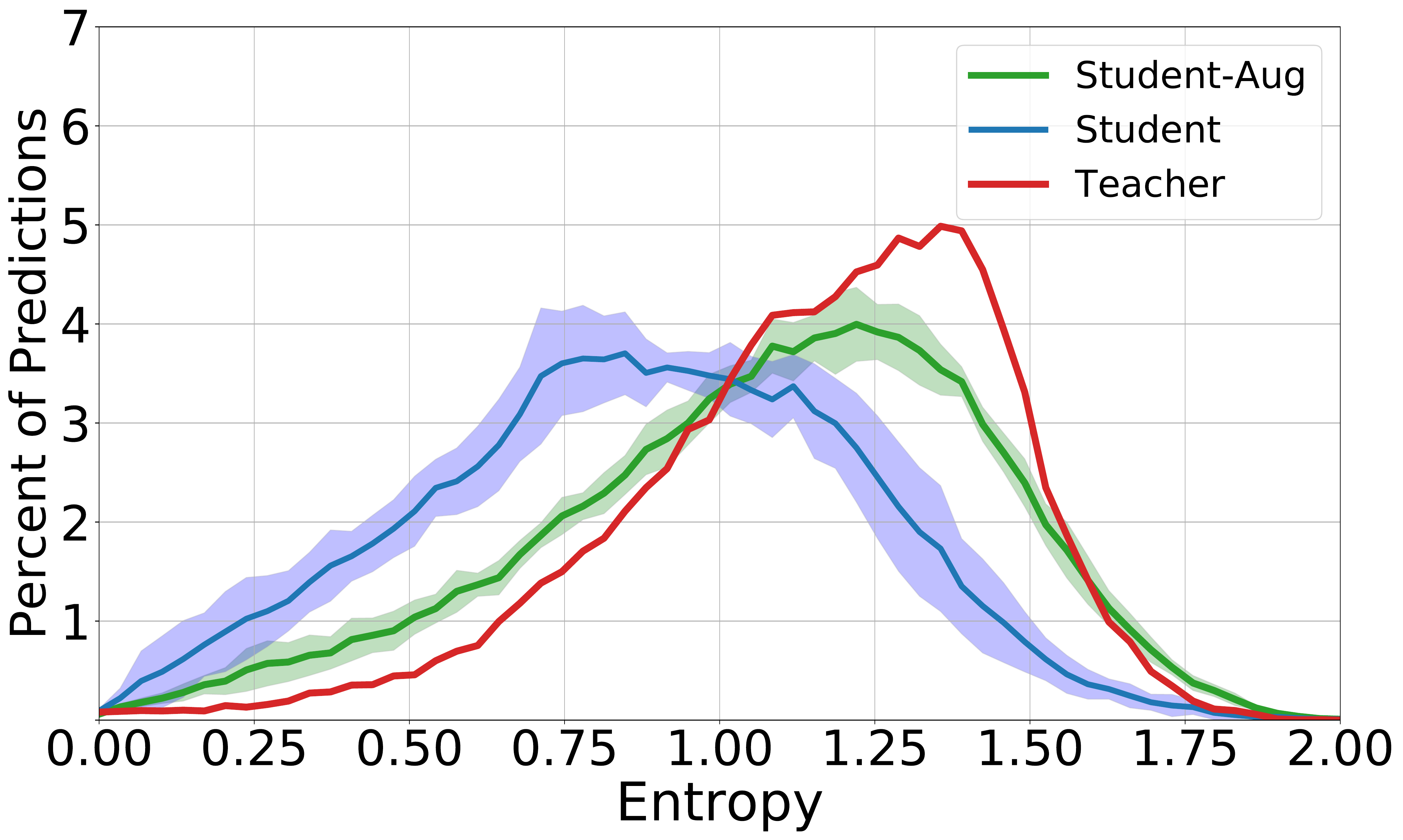} }}%
    \qquad
    \subfigure[CIFAR100-SVHN]{\label{fig:ood:cifar100}{\includegraphics[width=.3\linewidth]{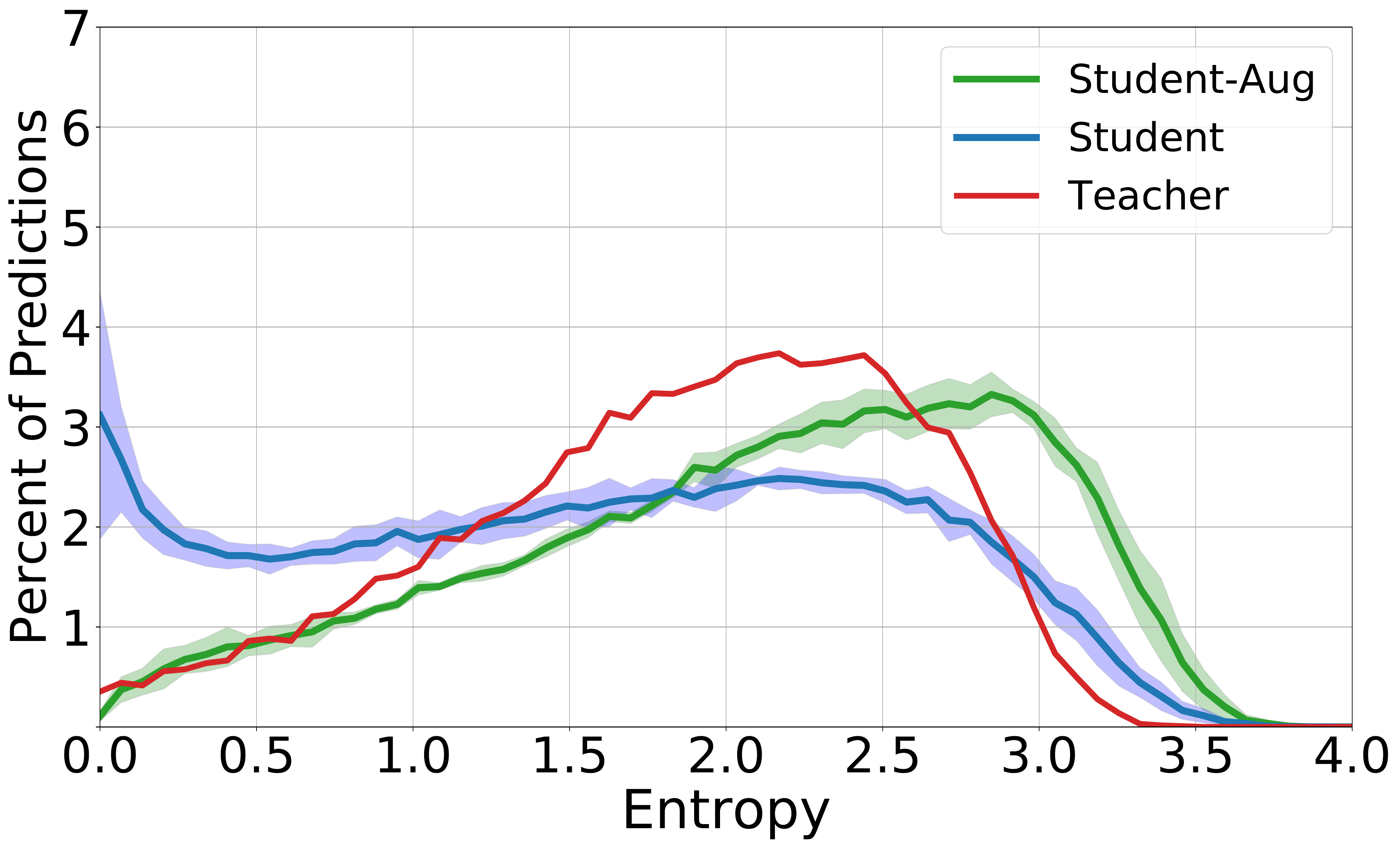}}}%
    \vspace{-0.3cm}
    \caption{\textbf{Out-of-Distribution:} Entropy histograms of the predictions of models trained on CIFAR datasets and evaluated on OOD SVHN dataset. Student corresponds to training with sharper targets through interpolation with true delta distribution, Student-Aug uses transformations to traverse the natural manifold. The teacher uses 15 networks of depth 5(a) or depth 18(b) and the students uses the same depth as their teacher. %Neither of the teacher and student use augmentation for MNIST.
    }%
    \vspace{-0.3cm}
    \label{fig:ood}%
\end{figure*}

% Maybe we should look at average entropy for the different targets too?

% TEXT: A more practical way of varying the target distribution is to decrease its entropy by performing a weighted average with the predictions of the teacher with the onehot target. ~\citep{hinton2015nips} motivates this by ....
%From KD: We have found that using the original training set workswell, especially if we add a small term to the objective function that encourages the small modelto predict the true targets as well as matching the soft targets provided by the cumbersome model.Typically, the small model cannot exactly match the soft targets and erring in the direction of thecorrect answer turns out to be helpful.

% TEXT: Figure/Table \ref{} shows that initially, increasing $\alpha$ improves performance until a certain sweet spot after which the performance degrades. 

% Relate this to ensembles of different capacities?
% Or is this explained by, soft targets proposes how important each prediction is, ie..e how hard we should work to make it large. If the capacity of the student is large enough then it might have been able to make more targets larger, and increasing alpha will do this...?

% TODO: Figure/Table - Increasing alpha and show sweet spot

% Run same alphas but with higher capacity? Show that sweet spot comes later or earlier?

\vspace{-0.2cm}
\subsection{Proper class-posterior distribution}

% TEXT: We observe that higher capacity students works better with smaller alpha.
% Low priority. Could have the sorted conf pictures here instead maybe?
% Capacity plot showing higher capacity needs lower alpha: DO THIS IF TIME
% --  alpha=0, alpha=0.05, alpha=0.1, alpha=0.2 for depth 5, 18 ?
% -- Have: a0d5, a0d18, a02d5
%    Need: a005d5, a005d18, a01d5, a01d18, a02d5, a02d18

% Connect to above if I have time to create figure.

As we discussed, a way to improve the distillation process is to correct for the wrong predictions of the teacher ensemble. We proposed another interpretation of the weighted average between the true label and the teacher predictions. We move $\alpha_i$ in Eq~\ref{eq:proper_target} in the range $\frac{p(y_{max})-p(y_i)}{p(y_{max})-p(y_i)+1}<\alpha_i\leq1$. We observed no significant difference when using the approach on CIFAR10, however, it gives small improvements on CIFAR100 (Tab~\ref{tab:proper_cifar100} in the appendix). We hypothesize the reason for this is that the number of misclassified examples was too low for CIFAR10 for the change to show significant improvement and thus going to a more challenging task such as ImageNet would further signify the benefits.
\vspace{-0.2cm}
\subsection{Robustness to out-of-distribution samples}
% we point to the first figure and say that while the in-dist results are great, the OOD results are far from ideal. So, ...
In Fig~\ref{fig:a0}, we saw that the uncertainty estimates for in-distribution samples are on par with the ensemble, especially with increased student's capacity while Fig~\ref{fig:vanilla} shows the robustness to OOD samples is far from ideal. We propose a simple approach for the natural OOD samples. Here we simply perturb the samples of the natural manifold by applying image transformations that do not violate the manifold including cropping and mirroring. 
In the standard case, the label for an augmented image is the teacher's prediction for the corresponding unperturbed image. We instead propose to use the teacher's prediction on the augmented image as the label, providing more information about the teacher during training.
%We instead propose that the label for an augmented image be the teacher's prediction for that image. Having the knowledge of what the teacher does for more than just the unperturbed training set gives more information for the student to use during training. 
Fig~\ref{fig:ood} shows the intriguing improvement this simple technique brings. The noticeable improvements we get from this simple approach, highlights the promise of pursuing this direction further.
%While this can be seen as a simple modification of the standard augmentation approach (used for student training), the noticeable improvements we get from this simple approach, highlights the promise of pursuing the general approach as outlined in Sec~\ref{sec:method:ood}. 
Interestingly, we have found that more aggressive transformations which is usually harmful for standard training helps the teacher-student learning.

% FOR NIPS: this most likely doesn't work because on the training set the number of wrongly classified samples by the teacher is very low

% Comment: Think we have to mention the difference between the standard and augmented somewhere in method
% 2.3 

% here we try to walk on the natural image manifold by the means of augmentation techniques that does not violate the natural manifold. The techniques we use include cropping, translation, mirroring. (FOR NIPS: correct the way we do the augmentation and expand on the techniques)

% current OOD Figure: with the introduction of this simple technique we can actually be on a par with the teachers on the NLL for OOD samples. probably discuss in detail different datasets.

% interestingly, this actually improves in-distribution results look at figure XX

% OOD figures
% Refer to figure 2 in appendix showing that aug helps for in-distribution too. and table for cifar100?

% NIPS: take the best approach, go to some high-profile challenge such as imagenet and produce state-of-the-art results! That would be super cool!

\vspace{-0.3cm}
\section{Final remarks}
\label{conclusion}
In this work we closely analyzed the distillation process of \cite{hinton2015nips} from an uncertainty estimation perspective. We shed light on their design choices which resulted into suggesting additional improvements. In the experimental part of this work we empirically studied the suggested aspects which led to many interesting observations. \\
Throughout all the experiments we tried to keep high-level experimental standard in our reports by cross-validated hyperparameter optimisation for baseline students. We also reported values as result of 5 different runs. %This prohibited us from exhaustively searching through all combinations of proposed techniques. However, we believe the theoretical and empirical observations will lead to fruitful discussions at the workshop.
Important future directions include theoretical analysis of our observations regarding the effects of distillation and its design choices and applying the techniques on larger datasets such as ImageNet to further highlight their effectiveness.
% - Give final results here for MNIST, CIFAR10, CIFAR100 and compare to ensemble. Say that the computational efficiency is hugely different.

% - review what we did and found.

% - Finally, with the proposed method we hold the state of the art on uncertainty estimation while remain faster than the previous methods.

% - pruning and compression techniques can always be used to make the student faster instead of crippling the student capacity for learning more complicated distributions.

\vspace{-0.1cm}
\section*{Acknowledgements}
This work was supported by the Wallenberg AI, Autonomous Systems and Software Program (WASP-AI/MLX) funded by the Knut and Alice Wallenberg Foundation.

%\textbf{Do not} include acknowledgements in the initial version of the paper submitted for blind review.

\clearpage

\nocite{langley00}

\bibliography{main}
\bibliographystyle{icml2019}

\clearpage

\appendix
\section{Experimental Details}
\label{sec:app:exp_details}
We evaluate our method on MNIST, CIFAR-10 and CIFAR-100 training a dense neural network on MNIST and ResNet variants for CIFAR. To simplify things, we only consider ensembles of networks where each network has the same capacity. 

All models are trained on a train-validation split, where hyperparameters are optimized on the validation set based on NLL. We do not retrain on the entire training set before evaluating on the test set. 
%early stopping for teachers, difference between teachers are seeing the data in different order during training because of shuffling and also different random initializations of the weights.

% Optimizer, hyper-parameters(learning rate, learning rate scheduling, epochs, bs, ) 
\textbf{MNIST} For all MNIST experiments, we use the same dense neural network architecture as proposed in ~\citep{lakshminarayanan2017nips}. That is,  three hidden layers with 200 units per layer, ReLU activations and batch normalization. Both the student and the teacher is trained using the Adam optimizer. Each network of the teacher is trained for 10 epochs with a fixed learning rate of 0.001 and a batch size of 1000. The students are trained for 600 epochs with a fixed learning rate of 0.002 and a batch size of 64.

\textbf{CIFAR} For all CIFAR-10 and CIFAR-100 experiments, we use the ResNet version proposed by ~\citep{he2016corr}. We use ResNet models of varying depth 5(ResNet32), 9(ResNet56), 18(ResNet110), etc. We train these models using the Momentum optimizer with a batch size of 128 and a learning rate of 0.1. The teacher networks overfit quickly to NLL after the first drop in learning rate. We drop the learning rate at epoch 82 and do early stopping at epoch 85 before the validation NLL degrades. The students can be trained for longer without overfitting to NLL. We use 2500 epochs and the learning rate is reduced by a factor of 10 at epoch 2000, 2100, 2300.  For data augmentation we use padding, random cropping and horizontal flips. As the baseline, the label for each augmented image is the prediction of the teacher on the corresponding original image. For the improved augmentation technique, the label for each augmented image is the prediction of the teacher for that particular augmented image.

% ---- In-Distribution Uncertainty -----
% -- Typical plots here: (err, nll, brier)-varying networks, (err, nll, brier)-varying everything, calibration plot, confsort -- 
% 1. Reducing parameters: Understanding the effect of the raw teacher predictions
%    - Varying depth of teacher, number of networks in teacher, types of teachers(MCDO etc) and depth of student
% 2. Comparing alpha and correct preds + alpha
%    - Would it be good to see the effect of varying alpha?
%    - Relate alpha to increasing number of networks in ensemble
% 3. Comparing effect of adding new class-specific labels(figure)

% Figure 2 for mnist and cifar10
\newpage
\section{Additional Figures and Tables}

% Same information in the figure below
\begin{table}[H]
\caption{\textbf{Sharpening the target distribution for CIFAR-10:} An ensemble of 15 teachers with depth 5 is distilled to a student of depth 18 with varying $\alpha$ in order to sharpen the target distributions. However, we see that as the student capacity is already addressing the regularization effect caused by the dispersed target distribution, the importance of $\alpha$ is diminished.}
\label{tab:sharpen}
\vskip 0.15in
\begin{center}
\begin{small}
\begin{sc}
\begin{tabular}{lcccc}
\toprule
$\bm{\alpha}$ & 0.0 & 0.1 & 0.2 & 0.3 \\
\midrule
$\textbf{NLL}_{mean}$ & 0.1573 & 0.1575 & 0.1575 & 0.1572 \\
$\textbf{NLL}_{std}$ & 0.0016 & 0.0016 & 0.0014 & 0.0012\\
\bottomrule
\end{tabular}
\end{sc}
\end{small}
\end{center}
\vskip -0.1in
\end{table}

%d5ds18 mean and std NLL from above:
%a0.0: 0.15733493199999998, 0.0015458989560304406
%a0.1: 0.15753434999999999, 0.0015774801698436656
%a0.2: 0.15756205333333331, 0.0013647934781578685
%a0.3: 0.1571644525, 0.0011883349277955037

\begin{table}[H]
\caption{\textbf{Proper class-posterior distribution for CIFAR-100:} An ensemble of 15 teachers with depth 5 is distilled to a student of depth 5. Here we correct the target distribution for the wrongly classified samples by the teacher ensemble. See Eq~\ref{eq:proper_target}. We vary $\alpha_i$ in the range  $\frac{p(y_{max})-p(y_i)}{p(y_{max})-p(y_i)+1}<\alpha_i\leq1$ which spans the spectrum of dark knowledge preservation constrained on the prediction (argmax) being correct. While the same experiment for CIFAR10 did not show significant improvement, we observe some promise moving to CIFAR100. We posit this is due to the fact that there are only a few wrongly classified samples in CIFAR10 compared to CIFAR100. Consequently, we expect larger improvements when going to more challenging tasks such as ImageNet. Here we denote the lower bound by $\check{\alpha_{i}}\triangleq\frac{p(y_{max})-p(y_i)}{p(y_{max})-p(y_i)+1}$}. ``Ref", refers to the baseline case where no $\alpha_i$ is used.
\label{tab:proper_cifar100}
\vskip 0.15in
\begin{center}
\begin{small}
\begin{sc}
\begin{tabular}{lccc}
\toprule
$\bm{\alpha}_i$ & $\check{\alpha_{i}}$ & $0.9\check{\alpha_{i}}+0.1$ & $0.8\check{\alpha_{i}}+0.2$\\
\midrule
$\textbf{NLL}_{mean}$ & \textbf{0.920} & 0.922 & 0.925\\
$\textbf{NLL}_{std}$  & 0.004          & 0.005 & 0.003\\
\midrule
$\textbf{Ref}_{mean}$ &  & 0.922 &\\
$\textbf{Ref}_{std}$ &  & 0.007 &\\
\bottomrule
\end{tabular}
\end{sc}
\end{small}
\end{center}
\vskip -0.1in
\end{table}

\begin{table*}[t]
\caption{\textbf{In-distribution performance on CIFAR-100:} An ensemble of 15 teacher networks of depth 18 is distilled to a student of depth 18. The classification error, NLL, and Brier score for the teacher(baseline) and students of varying depth and mixing parameter $\alpha$. Sharpening the targets($\alpha>0$) does not improve the performance, however increasing the depth of the student makes a significant improvement.}
\label{table:appendix:cifar100-sharpen}
\vskip 0.15in
\begin{center}
\begin{small}
\begin{sc}
\begin{tabular}{ccc|ccc}
\toprule
& Depth & $\alpha$ & Error & NLL & Brier Score \\
\midrule
Teacher & 18 & - & 20.95 & 0.7434 & 0.00296 \\
Student & 18    & 0.0 & 22.92 $\pm$ 0.15 & 0.8187 $\pm$ 0.0060 & 0.00320 $\pm$ 0.00002\\
Student & 18    & 0.1 & 23.24 $\pm$ 0.29 & 0.8257 $\pm$ 0.0049 & 0.00322 $\pm$ 0.00002\\
Student & 27    & 0.0 & 22.16 $\pm$ 0.15 & 0.7856 $\pm$ 0.0070 & 0.00310 $\pm$ 0.00003\\
Student & 27    & 0.1 & 22.26 $\pm$ 0.15 & 0.7923 $\pm$ 0.0044 & 0.00311 $\pm$ 0.00002\\
\bottomrule
\end{tabular}
\end{sc}
\end{small}
\end{center}
\vskip -0.1in
%----------------
% ---- Table CIFAR100 in-distribution using Augmentation ----

%\caption{\textbf{In-distribution performance on CIFAR-100:} The classification error, NLL, and Brier score for students of varying depth and mixing parameter $\alpha$. All students are trained on teachers that use 15 networks of depth 18. Here the augmentation approach where the student is trained on the predictions of the teacher on all augmented images is used.}\label{table:appendix:cifar100-fullaug}
%\vskip 0.15in
%\begin{center}
%\begin{small}
%\begin{sc}
%\begin{tabular}{cc|ccc}
%\toprule
%Student Depth & $\alpha$ & Error & NLL & Brier Score \\
%\midrule
%18    & 0.0 & 23.48 $\pm$ 0.28 & 0.8085 $\pm$ 0.0026 & 0.00324 $\pm$ 0.00001 \\
%18    & 0.1 & 23.26 $\pm$ 0.11 & 0.8027 $\pm$ 0.0028 & 0.00322 $\pm$ 0.00001 \\
%27    & 0.0 & 23.09 $\pm$ 0.15 & 0.7982 $\pm$ 0.0026 & 0.00319 $\pm$ 0.00001 \\
%27    & 0.1 & 22.80 $\pm$ 0.18 & 0.7913 $\pm$ 0.0047 & 0.00317 $\pm$ 0.00002 \\
%\bottomrule
%\end{tabular}
%\end{sc}
%\end{small}
%\end{center}
%\vskip -0.1in
\end{table*}

\begin{figure*}%
    \centering
    \subfigure[Depth 5]{\label{fig:varying-num-networks-d5}{\includegraphics[width=.31\linewidth]{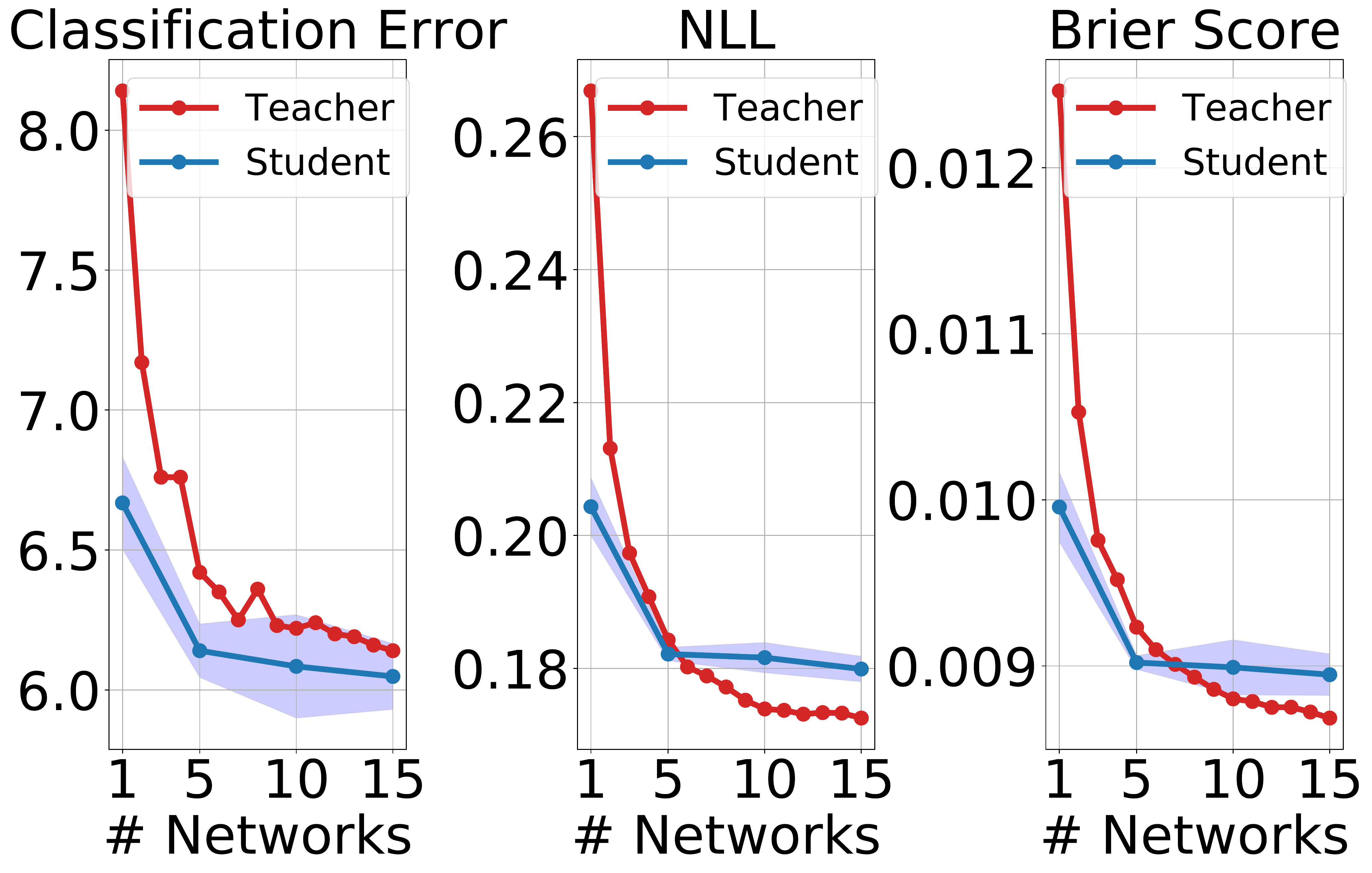}}}%
    \quad
    \subfigure[Depth 9]{\label{fig:varying-num-networks-d9}{\includegraphics[width=.31\linewidth]{Figures/cifar-figure2-onehot-single-depth9.pdf}}}%
    \quad
    \subfigure[Depth 18]{\label{fig:varying-num-networks:d18}{\includegraphics[width=.31\linewidth]{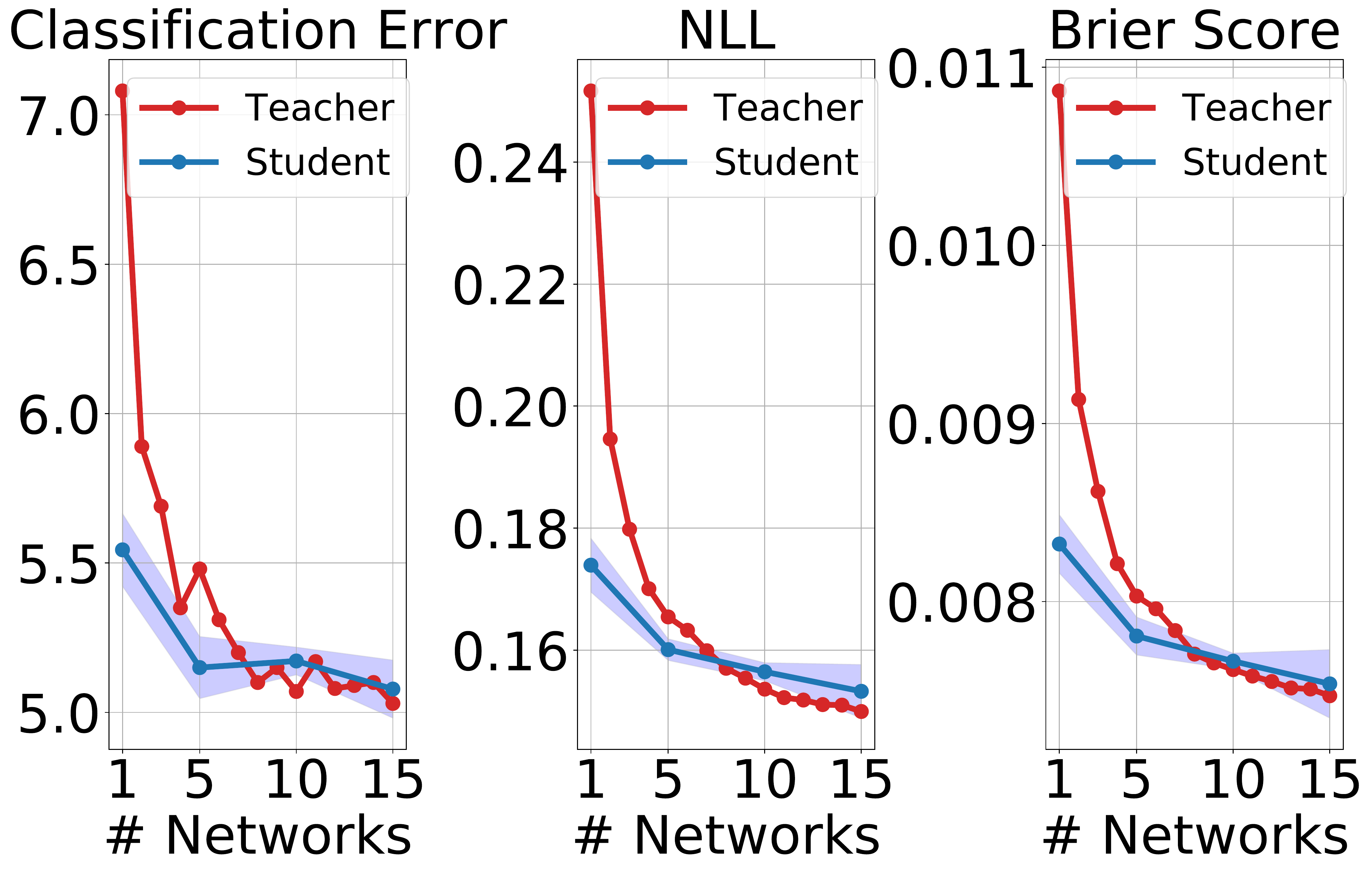}}}%
    
    \caption{\textbf{Vanilla distillation for different depths on CIFAR-10:} Evaluating the quality of the predictive uncertainty for the teacher and students trained on CIFAR-10 using vanilla distillation (equivalent to standard distillation with $\alpha=0, T=1$). The teacher and student is compared in terms of classification error, NLL and Brier score for varying depths. The teacher and student uses the same depth for their networks.}%
    \label{fig:varying-num-networks}%
\end{figure*}

% TODO: For NIPS -- Have a figure showing student depth fixed but vary capacity of teacher that generated the target. Think this will look like the standard loss curves for overfitting, i.e. bowl

%\begin{figure*}%
%    \centering
%    \subfigure[Train]{\label{fig:conf-sort:train}{\includegraphics[width=.45\linewidth, trim={0 0 0 %0},clip]{Figures/conf_sort_train_several.pdf}}}%
%    \subfigure[Test]{\label{fig:conf-sort:test}{\includegraphics[width=.45\linewidth, trim={0 0 0 0},clip]{Figures/conf_sort_test_several.pdf}}}%
%    \caption{\textbf{Confidence profile of different ensemble teachers:} Illustrating how the confidence of an ensemble of 15 networks(teacher) varies for different depths. Each teacher is evaluated on the a)train or b)test sets, the predictions are then sorted based on the confidence in the correct class. This sorting is done for each teacher independently.}%
%    \label{fig:conf-sort}%
%\end{figure*}

\begin{figure*}%
    \centering
    \subfigure[Teacher Depth 5]{\label{fig:appendix:a0:varying-student-depth:5}{\includegraphics[width=.3\linewidth, trim={0 0 0 0},clip]{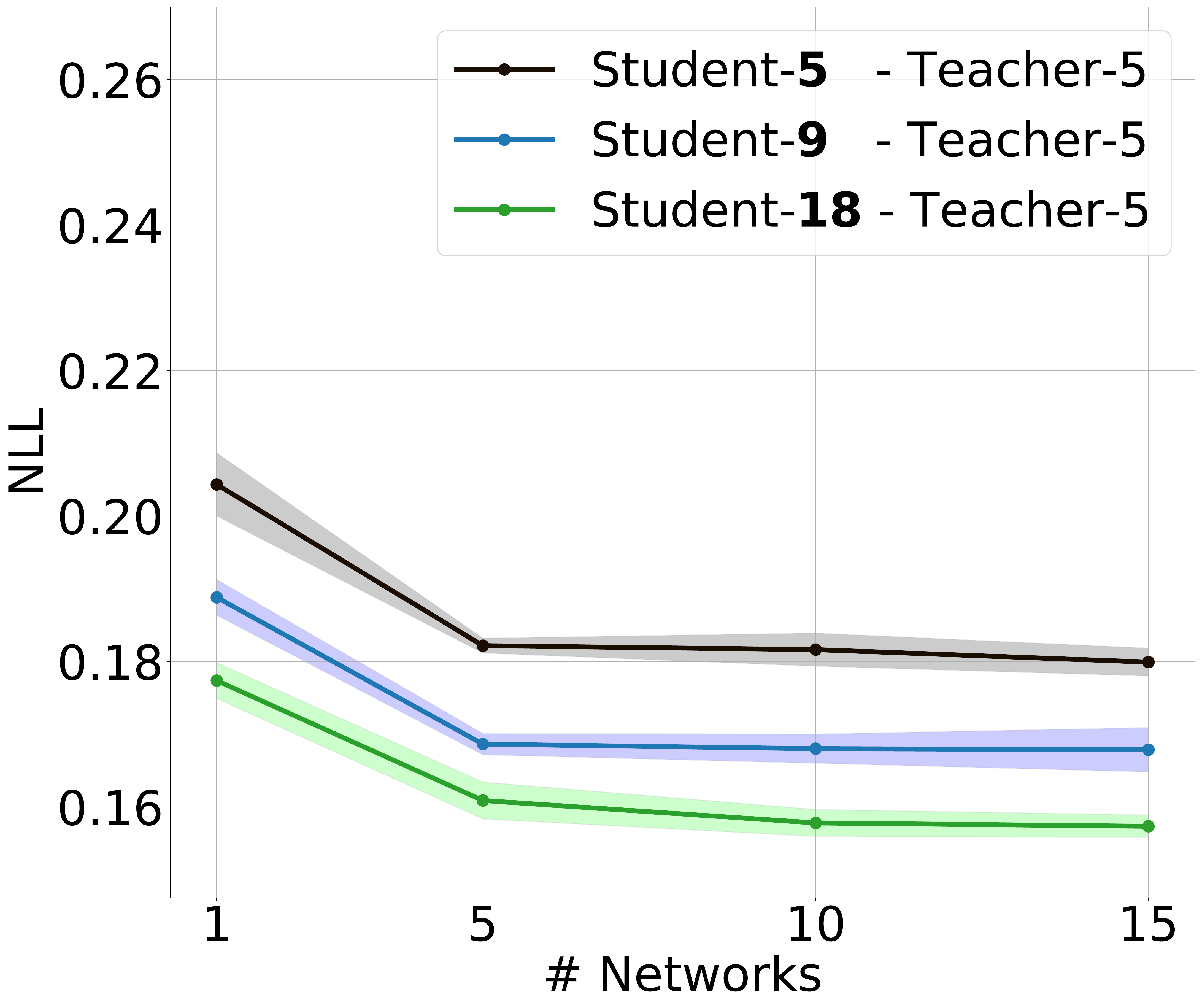}}}%
    \subfigure[Teacher Depth 9]{\label{fig:appendix:a0:varying-student-depth:9}{\includegraphics[width=.3\linewidth]{Figures/loaded-effect-of-teacher-networks-varyingstudentcapacity9.pdf}}}% 
    \subfigure[Teacher Depth 18]{\label{fig:appendix:a0:varying-student-depth:18}{\includegraphics[width=.3\linewidth]{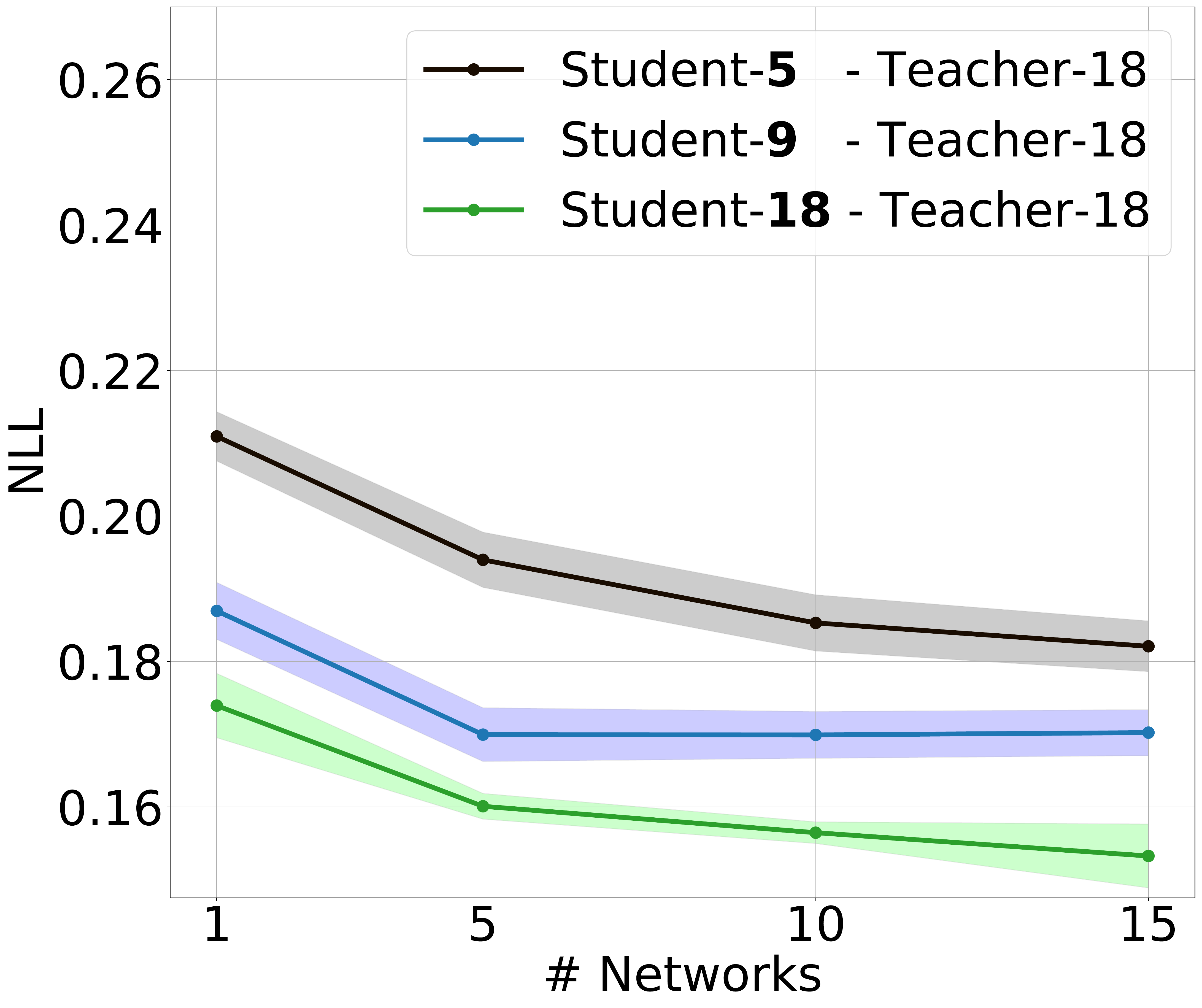}}}%
    %\subfloat[CIFAR100-SVHN]{{\includegraphics[width=.2\linewidth]{student-1-varing-capacity-networks.pdf}}}%
    %\hspace*{0em}
    %\vspace{-0.3cm}
    \caption{\textbf{Performance impact of student depth on CIFAR-10:} Increasing the depth of the student, while keeping the teacher depth fixed, leads to better NLL. This improvement is observed for teachers of depth 5(a), 9(b) and 18(c). Increasing the depth of the student consistently improves the result, no matter the number/depth of networks used by the teacher.}%
    %\vspace{-0.5cm}
    \label{fig:appendix:a0-varying-student-depth}%
\end{figure*}

\begin{figure*}%
    \centering
    \subfigure[Student Depth 5]{\label{fig:appendix:a0:varying-teacher-depth:5}{\includegraphics[width=.3\linewidth, trim={0 0 0 0},clip]{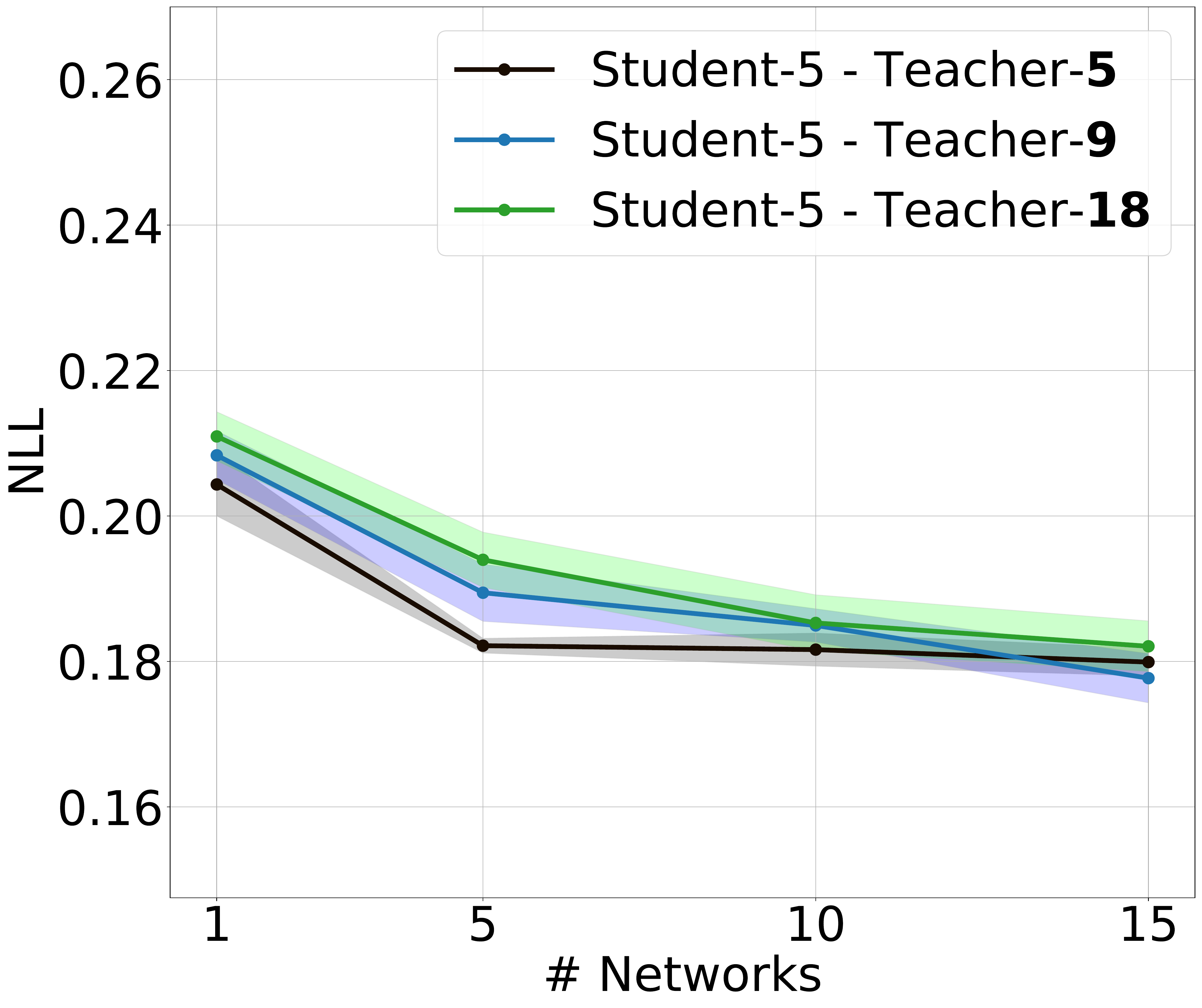}}}%
    \subfigure[Student Depth 9]{\label{fig:appendix:a0:varying-teacher-depth:9}{\includegraphics[width=.3\linewidth]{Figures/loaded-effect-of-teacher-networks-varyingteachercapacity-9.pdf}}}% 
    \subfigure[Student Depth 18]{\label{fig:appendix:a0:varying-teacher-depth:18}{\includegraphics[width=.3\linewidth]{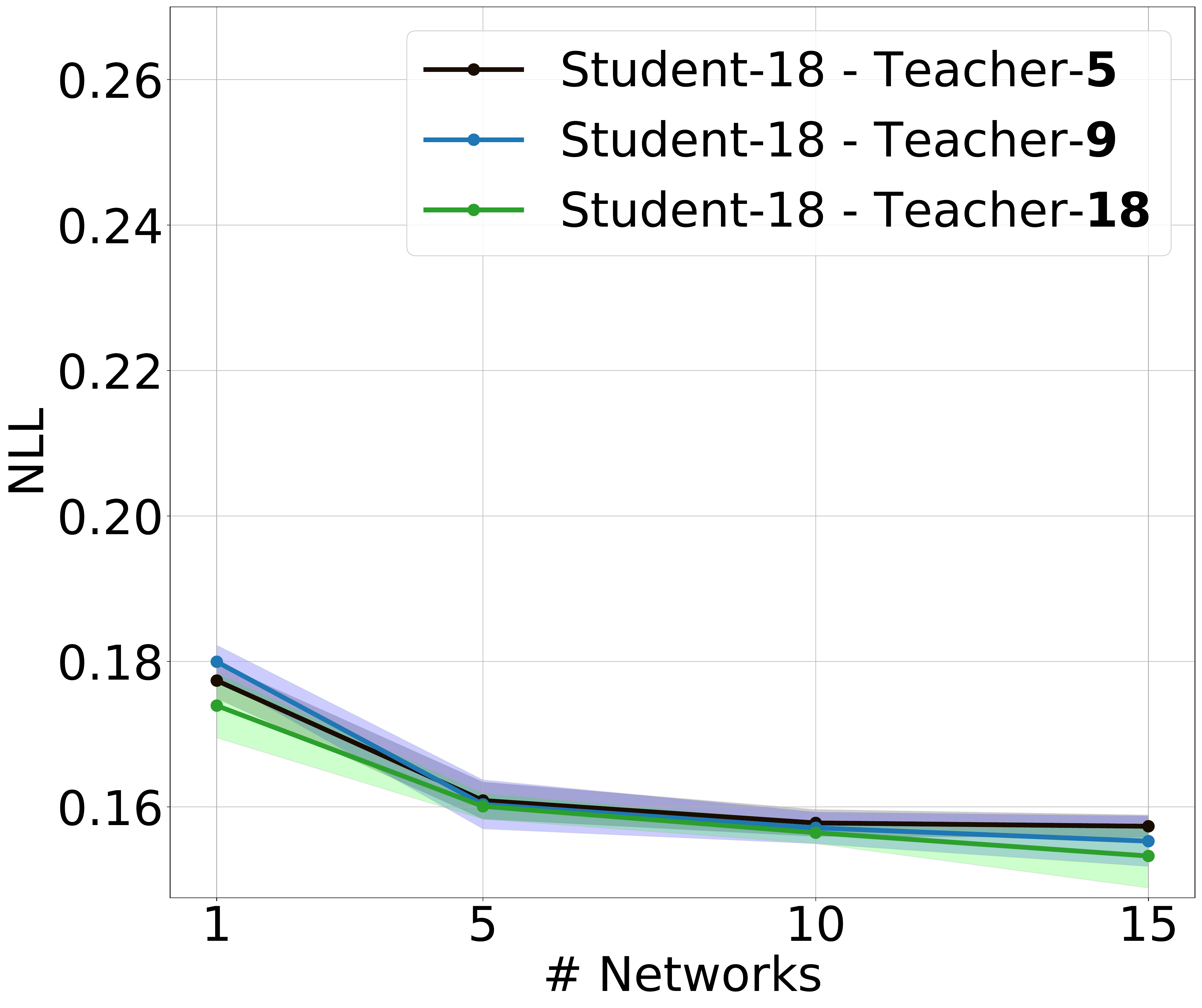}}}%
    %\subfloat[CIFAR100-SVHN]{{\includegraphics[width=.2\linewidth]{student-1-varing-capacity-networks.pdf}}}%
    %\hspace*{0em}
    %\vspace{-0.3cm}
    \caption{\textbf{Performance impact of teacher depth on CIFAR-10:} Varying the depth of the teacher, while keeping the depth of the student fixed, has no significant effect on the performance of the student. This seem to hold for students of different depths, see (a), (b) and (c). No matter the number of networks used by the teacher, or the depth of the student, varying the depth of the teacher does not significantly affect the student's performance.}%
    %\vspace{-0.5cm}
    \label{fig:appendix:a0-varying-teacher-depth}%
\end{figure*}

\begin{figure*}%
    \vspace{-1.5cm}
    \centering
    \subfigure{\label{fig:appendix:mnist}{\includegraphics[width=.4\linewidth]{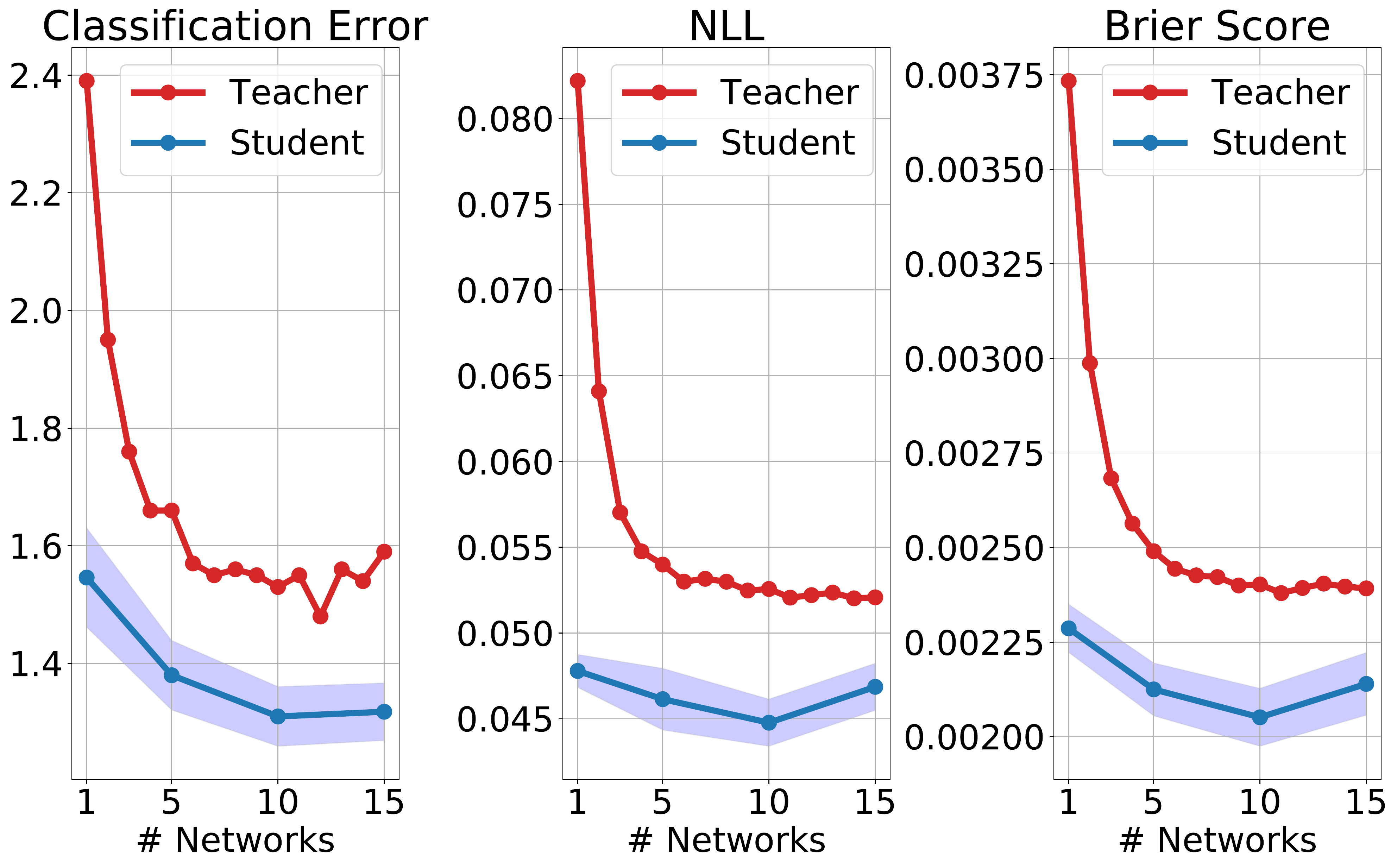}}}%
    %\qquad
    %\subfigure[CIFAR-10]{\label{fig:example:cifar10}{\includegraphics[width=.45\linewidth]{Figures/cifar-figure2-onehot-latest.pdf} }}%
    \caption{\textbf{In-distribution quality of uncertainty for MNIST:} Evaluating the quality of the predictive uncertainty for the teacher and students trained on MNIST using an alpha of 0.2. The performance is shown for varying number of networks. For the student, this corresponds to how many networks was used by its teacher.}%
    \label{fig:appendix:mnist}%
\end{figure*}

\end{document}